\documentclass{article}

\usepackage[dblblindworkshop, final]{neurips_2026} 

\usepackage[utf8]{inputenc} % allow utf-8 input
\usepackage[T1]{fontenc}    % use 8-bit T1 fonts
\usepackage{hyperref}       % hyperlinks
\usepackage{url}            % simple URL typesetting
\usepackage{booktabs}       % professional-quality tables
\usepackage{amsfonts}       % blackboard math symbols
\usepackage{nicefrac}       % compact symbols for 1/2, etc.
\usepackage{microtype}      % microtypography
\usepackage{xcolor}         % colors
\usepackage{listings}
\usepackage{amssymb}

\usepackage{pgfplots}
\pgfplotsset{compat=1.18}
\usepackage{subcaption}

\newcommand{\metrique}{RVS} 

\title{Beyond Accuracy: On Measuring Logical Compliance of Predictive Models}
\workshoptitle{TAE (Trust-AI-Eval): Can We Trust AI Evaluation?}

\author{%
  Guillaume Delplanque \\
  \texttt{guillaume.delplanque@inria.fr} \\
  \And
  Pierre Genevès \\
  \texttt{pierre.geneves@inria.fr} \\
  \And
  Nabil Layaïda \\
  \texttt{nadia.layaida@inria.fr} \\
}

\begin{document}

\maketitle

\begin{abstract}
  Machine learning models are predominantly evaluated through predictive performance metrics such as ranking quality, prediction error, or classification accuracy.
  While these metrics effectively quantify how closely predictions match the ground truth, they do not assess whether model outputs respect predefined logical or domain-specific constraints.
  In high-stakes applications, including healthcare, finance, and autonomous systems, logical consistency can be as critical as predictive accuracy, yet no standard metric captures this dimension.
  We introduce the \emph{Rule Violation Score (\metrique{})}, a complementary evaluation metric that quantifies the extent to which a predictive model respects a given set of logical rules, independently of predictive accuracy.
  \metrique{} treats hard rules (strict constraints) and soft rules (statistical regularities) differently, can be evaluated on any dataset and on any predictive model expressed over a relational vocabulary, and can be computed using SQL queries that are automatically generated for Horn rules.
  Beyond evaluating models, \metrique{} can also evaluate the logical consistency of training datasets and help identify poorly defined rules.
  We evaluate \metrique{} on three benchmarks covering knowledge graph link prediction and relational regression, including rule-based, embedding-based, and neuro-symbolic predictive models.
  Our results demonstrate that two models achieving comparable predictive accuracy can exhibit substantially different levels of logical compliance, revealing differences in model behavior that standard metrics fail to capture.

\end{abstract}

\section{Introduction}
\label{sec:intro}

Machine learning algorithms are commonly evaluated through predictive performance metrics: the central question is usually how closely their predictions match a reference ground truth. This focus extends even to advanced systems that explicitly exploit structured knowledge, including graph-based neuro-symbolic models \citep{delplanque-nesy25,nbfnet} and relational learning systems \citep{relbenchv2,relgnn}. These systems are still most often assessed through the prism of predictive accuracy, using task-specific metrics such as mean reciprocal rank (MRR) and Hits@$k$ \citep{hitsatk-metric} for link prediction, mean absolute error (MAE) \citep{mae-metric} and $R^2$ \citep{r2-metric,chicco2021coefficient} for regression, or accuracy and precision \citep{precision-metric} for classification \citep{sokolova2009systematic,powers2011evaluation}.

However, these metrics effectively quantify predictive performance, but they often fail to assess whether model outputs satisfy logical or domain-specific constraints. In many high-stakes applications, including healthcare, finance, cybersecurity or autonomous systems \cite{giunchiglia2023road}, respecting logical consistency and operational rules can be as important as, or even more important than maximizing predictive accuracy alone. A model achieving high accuracy may generate outputs that violate critical constraints, potentially leading to unreliable or unsafe decisions.

In this paper, we argue for an explicit measure of model compliance with logical constraints, and we introduce \emph{Rule Violation Score (\metrique{})}, an evaluation metric designed to quantify the extent to which a predictive model respects predefined logical rules, independently of its predictive accuracy. We built \metrique{} so that it satisfies several properties that we identify as desirable in a model evaluation protocol, especially for evaluating rapidly evolving state-of-the art algorithms for predicting links and node attributes in knowledge and relational databases.
Experiments show that two predictive models with similar accuracy scores can exhibit significantly different levels of \metrique{}, highlighting important behavioral differences that remain invisible when relying solely on conventional performance metrics. Our results suggest that \metrique{} provides complementary information to traditional evaluation metrics and should be considered when assessing AI systems deployed in constraint-sensitive environments.

\subsection{Motivating example}

Consider the dataset $\mathcal{D}$ consisting of the three facts shown in Table~\ref{tab:example} and the link-prediction query
{\small \texttt{wife(Alice, ?)}}, which asks for Alice's spouse.
Suppose that two predictive models, Model~1 and Model~2, each return a single candidate completion for this query.

\begin{table}[h]
\begin{small}
\centering
\begin{tabular}{ll}
\hline
Known facts ($\mathcal{D}$) &
\begin{tabular}[t]{l}
{\small \texttt{mother(Alice, Bob)}}\\
{\small \texttt{father(Christopher, Bob)}}\\
{\small \texttt{sister(Alice, Daniel)}}
\end{tabular}
\\
\hline
Ground-truth answer &
{\small \texttt{wife(Alice, Emmanuel)}}
\\
\hline
Prediction of Model~1 &
{\small \texttt{wife(Alice, Christopher)}}
\\
Prediction of Model~2 &
{\small \texttt{wife(Alice, Daniel)}}
\\
\hline
\end{tabular}
\caption{Motivating example}\label{tab:example}
\end{small}
\end{table}

Both predictions are incorrect according to standard metrics.
However, Model~2 violates the hard constraint $\bot \leftarrow \texttt{sister}(x,y) \land \texttt{wife}(x,y),$ which states that two individuals cannot simultaneously be siblings and spouses, and is
assumed to admit no exception.
Model~1, by contrast, remains logically consistent.
Existing metrics do not capture this distinction.
This example illustrates that some predictions should be penalized not only for being incorrect but also for violating symbolic domain knowledge.
It motivates the need for evaluation metrics that complement accuracy by measuring the extent to which predictions respect domain knowledge.

\subsection{Desirable properties} \label{sec:properties}

For a compliance metric to be genuinely adoptable in benchmarks, we identify several properties.

\begin{description}
\item[P1 -- Ground-truth independence.] Can be computed without access to ground-truth labels.
\item[P2 -- Hard/soft rule distinction.] Hard rules are strict constraints (e.g.\ $\bot \leftarrow \mathrm{sister}(x,y)\wedge\mathrm{wife}(x,y)$); soft rules are statistical regularities admitting exceptions (e.g.\ $\mathrm{wife}(x,z)\leftarrow \mathrm{mother}(x,y)\wedge\mathrm{father}(z,y)$). Violations of each must be scored differently.
\item[P3 -- Direct interpretability.] Output as simple counts and ratios, not an opaquely weighted formula, so a non-expert reading a benchmark table understands a value without further explanation — and so rule-level or dataset-level diagnostics stay legible on their own.
\item[P4 -- User-controllable rule importance.] Domain experts can assign an explicit, settable importance weight per rule, used at aggregation, to inject knowledge the data alone cannot reveal.
\item[P5 -- Dataset-derived rule importance (with user override).] When hand-weighting hundreds of mined rules is impractical, rule importance should default to a signal derived from the dataset itself, while remaining overridable by domain experts.  Dataset evidence can also provide a default hard/soft classification when no expert-defined type is available.  A model should not be penalized identically for contradicting a rule the data itself rarely satisfies versus one it almost always satisfies.
\item[P6 -- Relational scope.] Defined directly over relational vocabularies (knowledge graphs and relational databases), natively supporting link prediction and node-level prediction.
\item[P7 -- Model- and data-agnostic computability.] Automatically computable by a generic implementation given any facts, predictions, and rules, not tied to a specific algorithm or approach.
\end{description}

These properties define the design requirements for the metric developed in this paper.
Our objective is twofold: to define a compliance metric satisfying these properties,
and to assess whether it reveals information about model behavior that is not captured
by conventional predictive-performance metrics.

\section{Preliminaries}
 
\label{sec:preliminaries}

We first introduce the formal notions used throughout the paper, starting with the general evaluation setting on which the proposed metric is defined.

\paragraph{Evaluation setting.}
An evaluation setting is a triple
$
\mathcal E = (D,\widehat D,R)
$
where $D$ is a finite set of observed facts, $\widehat D$ is a finite set of predictions produced by a model, and $R$ is a finite set of logical rules whose compliance is evaluated.

\paragraph{Data model.}
$D,\hat D,R$ share a relational vocabulary $P$ of predicate symbols with fixed arities. Variables come from $V$, constants from domain $C=\mathrm{Const}(D\cup\hat D\cup R)$. An atom is $p(t_1,\dots,t_n)$ with $p^{(n)}\in P$; a ground atom (fact) has all terms in $C$. This covers link prediction ($\mathrm{wife(Alice,Christopher)}$), node classification ($\mathrm{label}(v,c)$), and attribute regression ($\mathrm{price}(h,250)$).

\textbf{Logical rules.} A rule $r:\psi_r\leftarrow\varphi_r$ has body $\varphi_r$ and head $\psi_r$, Boolean formulas over $P$ possibly with numerical comparisons, arithmetic/aggregate expressions, and decidable predicates. $V_{\psi_r}\subseteq V_{\varphi_r}=:V_r$. A \emph{positive Horn rule} has $\varphi_r=B_1\wedge\cdots\wedge B_m$, $\psi_r=H$, atoms $B_i,H$.

\textbf{Groundings.} A grounding is $\theta:V_r\to C$; $\Theta_r$ is the set of all such. For facts $F$ and formula $\varphi$, $F\models\varphi\theta$ denotes truth of the grounded formula (membership for atoms). $B^F_r=\{\theta\in\Theta_r\mid F\models\varphi_r\theta\}$.

\textbf{Body evaluability.} Every rule body must be evaluable over any finite $F$ in finite time, so $B_r^F$ is well-defined and computable.

\textbf{Confirmed/refuted heads.} For $\theta\in B_r^F$, $\psi_r\theta$ may be confirmed, refuted, or undecided. $C_r^F=\{\theta\in B_r^F\mid \psi_r\theta \texttt{ confirmed}\}$, $V_r^F=\{\theta\in B_r^F\mid \psi_r\theta \texttt{ refuted}\}$, $E_r^F=C_r^F\cup V_r^F$.

\textbf{Refutation mechanisms.} Under closed-world assumptions, absence refutes; under open-world assumptions, refutation needs explicit negative facts, incompatibility constraints, or decidable predicates. An incompatibility constraint $\bot\leftarrow\mathrm{brother}(x,z)\wedge\mathrm{aunt}(x,z)$ refutes $\mathrm{brother}(x,z)$ whenever $\mathrm{aunt}(x,z)$ holds. $R_{\mathrm{ref}}$ denotes such auxiliary rules, each of form $\bot\leftarrow\psi_r\wedge B_1\wedge\cdots\wedge B_m$.

\section{Rule Violation Score (\metrique) Definition}

Let $\mathcal E=(D,\widehat D,R)$ be an evaluation setting. For each $r\in R$, RVS computes
$d_r$, the contradiction rate in $D$, and $p_r$, the prediction-induced
contradiction rate relative to $D$.

\subsection{Dataset contradiction rate}
For a rule $r \in R$, the dataset contradiction rate $d_r$ measures the fraction of body-satisfying groundings in $D$ that lead to an established contradiction:
\[
d_r=\frac{|V_r^D|}{|E_r^D|}
\]
whenever $|E_r^D|>0$; otherwise $d_r$ is undefined.

$d_r$ is the default dataset-derived importance signal used to partition the rule set into hard and soft rules.
By default, $r$ is hard iff $d_r=0$ and soft otherwise, although users may explicitly declare a rule hard despite observed contradictions.
Let $R_{\mathrm{hard}}$ and $R_{\mathrm{soft}}$ denote the corresponding sets.

\subsection{Prediction contradiction rate}

Predictions are evaluated individually, via $D_A=D\cup\{A\}$ for each $A\in\widehat D$, so that contradictions can be associated with specific predictions.
For a grounded rule instance $r\theta$, let $\mathrm{Atoms}(r\theta)=\mathrm{Atoms}(\phi_r\theta)\cup\mathrm{Atoms}(\psi_r\theta)$ denote the ground atoms occurring in $\phi_r\theta$ or $\psi_r\theta$.
Prediction $A$ participates in $\theta$ iff $A\in\mathrm{Atoms}(r\theta)$.
We define:
$$
B_r^{D,A} = \{\theta\in\Theta_r \mid D_A\models\phi_r\theta,\; A\in\mathrm{Atoms}(r\theta)\},
$$
$$
C_r^{D,A} = \{\theta\in B_r^{D,A} \mid \psi_r\theta \texttt{ confirmed in }D_A\},
$$
$$
V_r^{D,A} = \{\theta\in B_r^{D,A} \mid \psi_r\theta \texttt{ refuted in }D_A\},
$$
$$
E_r^{D,A} = C_r^{D,A}\cup V_r^{D,A}.
$$

The prediction contradiction rate $p_r$ is defined as follows whenever the denominator is non-zero; otherwise $p_r$ is undefined:
\[
p_r=
\frac{\sum_{A\in\widehat D}|V_r^{D,A}|}
{\sum_{A\in\widehat D}|E_r^{D,A}|}
\]

\subsection{Rule-level score}
\label{sec:dr_pr_score}

For a hard rule $r\in R_{\mathrm{hard}}$, we define its rule-level score as the
number of prediction-induced contradictions,
\[
\mathbf{RVS}_r^{\mathrm{hard}}
=
\sum_{A\in\widehat D}|V_r^{D,A}|,
\]
with an optimal value of $0$. For a soft rule $r\in R_{\mathrm{soft}}$, the score compares the prediction contradiction rate with the dataset baseline:
\[
\mathbf{RVS}_r^{\mathrm{soft}}=\frac{p_r}{d_r}.
\]
A value below $1$ indicates better compliance than the dataset baseline, $1$ indicates
the same contradiction rate, and a value above $1$ indicates worse compliance. For
example, if $d_r=0.01$ and $p_r=0.03$, then
$\mathbf{RVS}_r^{\mathrm{soft}}=3$, meaning that predictions contradict the rule
three times as often as the dataset.

\subsection{Global aggregation}
\label{sec:global_metric_aggregation}
\paragraph{Hard rules.}
Given user-defined importance weights $\sigma_r>=0$ (P4), default $\sigma_r=1$:
\[
\mathbf{RVS}^{\mathrm{hard}}
=
\frac{\sum_{r\in R_{\mathrm{hard}}}
\sigma_r \cdot \mathbf{RVS}_r^{\mathrm{hard}}}{\sum_{r\in R_{\mathrm{hard}}} \sigma_r},
\]
whenever $\sum_{r\in R_{\mathrm{hard}}} \sigma_r>0$, otherwise $\mathbf{RVS}^{\mathrm{hard}}$ is undefined.
\paragraph{Soft rules.}
We define $W_r$ as a normalization factor corresponding to the fraction of
evaluable prediction-induced groundings associated with rule $r$.
Users may override these default weights. 
The aggregate soft-rule score is then defined as:
$$
W_r
=
\frac{\sum_{A\in\widehat D}|E_r^{D,A}|}
{\sum_{r'\in R_{\mathrm{soft}}}\sum_{A\in\widehat D}|E_{r'}^{D,A}|}, ~ ~ 
\mathbf{RVS}^{\mathrm{soft}}
=
\sum_{r\in R_{\mathrm{soft}}}
W_r \cdot \mathbf{RVS}_r^{\mathrm{soft}}
$$
\noindent Values below $1$ indicate better compliance than the dataset, values above $1$ indicate worse compliance.

% ============================================================
\subsection{RVS computation}
\label{sec:metric-computation}
% ============================================================

For positive Horn rules, SQL generation over a database storing $D,\hat D$ (one table per predicate) is fully automatic: body atoms compile into joins, shared variables into equality constraints, head confirmation/refutation into lookups or incompatibility checks. Non-Horn rules use rule-specific templates (Appendix~\ref{apd:computation}).

% ============================================================
\subsection{Verification of properties}
% ============================================================

 \metrique{} satisfies the properties introduced in Section~\ref{sec:properties}. Intuitively, for a hard rule, \metrique{} measures how many times the predictions contradict the rule, whereas for a soft rule, it measures how much the predictions contradict the rule relative to how often the same rule is contradicted in the dataset. Properties \textbf{(P1)} and \textbf{(P2)} -- ground-truth independence and the distinction between hard and soft rules -- thus hold immediately by construction. Direct interpretability \textbf{(P3)} follows because every reported quantity -- $d_r$, $p_r$, $\mathrm{RVS}^{\mathrm{hard}}_r$, $\mathrm{RVS}^{\mathrm{soft}}_r$, and their aggregates -- is a scalar computed from a single evaluation setting and is either a raw count, a ratio of counts, or a ratio of rates, with aggregation performed by a weighted mean. Consequently, the resulting rule-level and dataset-level diagnostics can be interpreted directly by practitioners without specialized knowledge of the metric; for instance, the value of $d_r$ alone can reveal whether a rule is poorly calibrated to the dataset. Rule importance is supported at both the user and dataset levels: user-controllable weights \textbf{(P4)} are specified directly through $\sigma_r$ for hard rules and through an optional override of $W_r$ for soft rules, while dataset-derived importance \textbf{(P5)} is reflected in the role of $d_r$ in determining the hard/soft distinction and in the default definition of $W_r$, which weights each soft rule according to its evidential support in the data. $d_r$ provides a rule-specific baseline for how frequently the rule is contradicted in the observed data. Thus, $\mathrm{RVS}^{\mathrm{soft}}_r=0$ when predictions never contradict the rule, $\mathrm{RVS}^{\mathrm{soft}}_r=1$ when they contradict it at the same rate as the dataset, and values respectively above or below $1$ indicate worse or better compliance than the dataset baseline. Moreover, for two models evaluated on the same rule with non-zero scores, the ratio of their \metrique{} scores reflects the ratio of their prediction contradiction rates relative to the same dataset baseline, thereby enabling interpretable model comparisons at the rule level. \metrique{} also satisfies relational scope \textbf{(P6)}, including link prediction, because $D$, $\hat{D}$, and $R$ are defined directly over a relational vocabulary without further assumptions. Finally, model- and data-agnostic computability \textbf{(P7)} follows because \metrique{} depends only on the evaluation setting and can therefore be applied to any predictive system. All these properties make \metrique{} suitable for use
alongside conventional predictive-performance metrics in benchmark evaluations:
all required quantities can be computed from $D$, $\hat{D}$, and $R$, independently of the predictive model being evaluated.

\section{Experiments}
\label{sec:exp}

We empirically evaluate whether RVS complements conventional predictive-performance
metrics across different tasks, data models, and predictive approaches.

\paragraph{Datasets and rules.} Table~\ref{tab:datasets} summarizes three datasets differing in domain, scale, and task.

\begin{table}[h]
\centering\small
\begin{tabular}{lrrrrrl}
\toprule
Dataset & Entities & Rel./Tables & Triples/Tuples & Test facts & Rules & Kind\\
\midrule
Family & 3{,}007 & 12 & 19{,}845 & 5{,}681 & 37 & Knowledge graph\\
FB15k-237 & 14{,}541 & 237 & 272{,}115 & 20{,}466 & 509 & Knowledge graph\\
DV3F & 392{,}809 & 13 & 4{,}837{,}087 & 2{,}213 & 4 & Relational database\\
\bottomrule
\end{tabular}
\caption{Dataset characteristics.}
\label{tab:datasets}
\end{table}

\emph{Family} \cite{family_base}: 36 AMIE+ rules \cite{Amie+} (e.g.\ $\mathrm{aunt}(A,C)\leftarrow\mathrm{sister}(A,B)\wedge\mathrm{father}(B,C)$) plus one deliberately imperfect soft rule $\mathrm{father}(X,Y)\leftarrow\mathrm{son}(Y,X)$ (37 total), plus incompatibility constraints (e.g.\ $\bot\leftarrow\mathrm{brother}(A,B)\wedge\mathrm{sister}(A,B)$). Strong semantic constraints over kinship; tests whether RVS detects implausible predictions even under high accuracy.

\emph{FB15k-237} \cite{FB15k237} (from Freebase \cite{Freebase}): 509 NeuralLP rules \cite{NeuralLP}, mostly soft, e.g.\ $\mathrm{teamLocation}(X,Y)\leftarrow\mathrm{placeOfBirth}(Y,X)$, plus automatically generated incompatibility constraints (Appendix~\ref{apd:incompatibility}). Larger, weaker regularities; tests RVS where rules are mostly statistical.

\emph{DV3F} \cite{DV3F}: French real-estate transaction-price regression. Proximity rules state comparable price/m\textsuperscript{2} for properties sharing type, postal code, and year, e.g.\ for houses: $|\mathrm{psm}(A)-\mathrm{psm}(B)|<\epsilon \leftarrow \mathrm{type}(A){=}\mathrm{type}(B){=}\texttt{``house''}\wedge \mathrm{pc}(A){=}\mathrm{pc}(B)\wedge \mathrm{year}(A){=}\mathrm{year}(B)$.

\paragraph{Predictive models.} AnyBURL \cite{anyburl} (rule-based); CompGCN \cite{compgcn}, GraphSAGE \cite{graphsage} (neural graph); UniKER \cite{uniker}, ExpressGNN \cite{ExpressGNN} (neuro-symbolic); Rel-LLM \cite{relllm} (GNN+LLM). Table~\ref{tab:models} summarizes coverage; $\hat D$ is obtained by running each model on the test partition.

\begin{table}[h]
\centering\small
\begin{tabular}{llll}
\toprule
Algorithm & Family & Approach & Datasets\\
\midrule
AnyBURL & Neuro-symbolic & Rule-based & Family, FB15k-237\\
CompGCN & Neural & KG embedding & Family, FB15k-237\\
UniKER & Neuro-symbolic & Embedding + rules & Family\\
ExpressGNN & Neuro-symbolic & GNN + prob.\ logic & FB15k-237\\
GraphSAGE & Neural & GNN & DV3F\\
Rel-LLM & Neural & GNN + LLM & DV3F\\
\bottomrule
\end{tabular}
\caption{Predictive models evaluated.}
\label{tab:models}
\vspace{-0.1cm}
\end{table}

\paragraph{Results.}
\emph{Family.} 
Figure~\ref{fig:family} (left) reports the rule-level contradiction rate $d_r$ computed on the \textsc{Family} dataset.
Most rules exhibit near-zero values, indicating that the dataset is largely consistent with the family relation semantics encoded in the rule set.
The main exception is the deliberately imperfect soft rule
$\texttt{father}(X,Y) \leftarrow \texttt{son}(Y,X)$,
for which $d_r = 0.467$. This confirms that the rule is contradicted in approximately $46.7\%$ of its evaluable groundings, as expected: observing
$\texttt{son}(Y,X)$ does not uniquely determine that $X$ is the father of
$Y$, since $X$ may also be the mother.
This illustrates the role of $d_r$ as a rule-level diagnostic metric: rules that are semantically too broad or poorly calibrated are immediately identifiable before evaluating any predictive model.

Table~\ref{tab:family} reports the average of Hits@10, MRR, $\mathbf{\metrique}^{\texttt{soft}}$, and $\mathbf{\metrique}^{\texttt{hard}}$ on 5 runs of each method.
AnyBURL achieves both the highest predictive performance (MRR $=0.973$ and Hits@10 $=0.984$) and the highest logical compliance ($\mathbf{\metrique}^{\texttt{hard}}=5.89$ and $\mathbf{\metrique}^{\texttt{soft}}=2.24$).
This is expected given its rule-based nature.

The comparison between CompGCN and UniKER is more informative.
CompGCN achieves a higher MRR than UniKER (MRR: $0.912$ vs.\ $0.874$), yet it produces substantially more hard-rule violations ($426$ vs.\ $62.5$).
Thus, the model with higher ranking performance is not the one that best satisfies hard logical constraints.
RVS therefore exposes a difference in model behavior that is invisible from MRR or Hits@10.

This effect is strongly dependent on the rule considered. This is illustrated in Figure~\ref{fig:family} (right), which reports the per-rule $\mathbf{\metrique}_r^{\texttt{hard}}$ values for CompGCN and UniKER on one randomly selected run of each algorithm. For example, on the hard rule
$\texttt{brother}(X,Z) \leftarrow \texttt{brother}(X,Y) \land \texttt{sister}(Y,Z)$, CompGCN produces $1160$ violations out of $4626$ evaluable prediction-induced groundings, whereas UniKER produces only $35$ out of $3158$. Conversely, for the rule
$\texttt{aunt}(X,Z) \leftarrow \texttt{sister}(X,Y) \land \texttt{father}(Y,Z)$,
CompGCN produces only $6$ violations out of $662$, whereas UniKER produces $27$ out of $1013$. This shows that CompGCN exhibits large variation across hard rules, while UniKER displays a more homogeneous violation profile. This per-rule view is informative: the RVS score summarises logical compliance, and rule-level scores indicate which logical behaviors are responsible.

\begin{table}[h]
\centering\small
\begin{tabular}{lrrrr}
\toprule
Method & Hits@10 & MRR & $\mathrm{RVS}^{\mathrm{soft}}$ & $\mathrm{RVS}^{\mathrm{hard}}$\\
\midrule
AnyBURL & \textbf{0.984 (0.000)} & \textbf{0.973 (0.002)} & \textbf{2.24 (0.15)} & \textbf{5.89 (0.27)}\\
CompGCN & 0.978 (0.004) & 0.912 (0.021) & 38.1 (3.4) & 426 (4.7)\\
UniKER & 0.978 (0.000) & 0.874 (0.006) & 16.3 (0.68) & 62.5 (2.8)\\
\bottomrule
\end{tabular}
\caption{Results on Family on 5 runs. Bold = best per metric. standard deviation in brackets.}
\label{tab:family}
\end{table}

\begin{figure}[ht]
\centering
\hspace{-2.5cm}
\vspace{-0.1cm}
% --- FIRST FIGURE ---
\begin{subfigure}{0.48\textwidth}
\centering
\begin{tikzpicture}
\begin{axis}[
    width=6.5cm,
    height=4.3cm,
    ymin=-0.01,
    ymax=0.103,
    xmin=0.5,
    xmax=37.5,
    xlabel={Rule index},
    ylabel={$d_r$},
    xtick={1,10,20,30,37},
    ytick={0,0.05,0.1},
    ymajorgrids,
    bar width=3pt,
]

% Rules with d_r = 0 (drawn in red)
\addplot[ybar, fill=red!60, draw=red!80] coordinates {
(3,0.0) (4,0.0) (6,0.0) (7,0.0) (8,0.0) (9,0.0) (10,0.0)
(11,0.0) (12,0.0) (13,0.0) (14,0.0) (15,0.0) (16,0.0) (17,0.0)
(18,0.0) (19,0.0) (21,0.0) (22,0.0) (23,0.0) (24,0.0) (25,0.0)
(27,0.0) (28,0.0) (29,0.0) (31,0.0) (32,0.0) (33,0.0) (34,0.0)
};

% Rules with d_r > 0 (drawn in blue)
\addplot[ybar, fill=blue!70, draw=blue!90] coordinates {
(1,0.0053) (2,0.0062) (5,0.0031) (20,0.0025) (26,0.0028)
(30,0.0067) (35,0.0036) (36,0.0054) (37,0.4667)
};

% Break mark + label for the outlier at rule 37
\node[left] at (axis cs:37,0.086) {0.47};
\draw (axis cs:36,0.09) -- (axis cs:36.5,0.103);
\draw (axis cs:37,0.103) -- (axis cs:37.5,0.09);

\end{axis}
\end{tikzpicture}
\end{subfigure}
\hspace{0.3cm}
% --- SECOND FIGURE ---
\begin{subfigure}{0.48\textwidth}
\centering
\begin{tikzpicture}
\begin{axis}[
    width=8cm,
    height=4.3cm,
    ymin=0,
    ymax=2100,
    xmin=0.5,
    xmax=28.5,
    xlabel={Rule index},
    ylabel={RVS},
    xtick={1,10,20,28},
    ytick={62.5,425,1000,2000},
    ymajorgrids,
    bar width=2pt,
    legend style={at={(0.6,0.95)}, anchor=north west, font=\small},
    area legend,
]

% Average reference lines
\addplot[orange!80, thick, dashed, forget plot] coordinates {(0.5,425) (28.5,425)};
\addplot[green!70!black, thick, dashed, forget plot] coordinates {(0.5,62.5) (28.5,62.5)};

% CompGCN
\addplot[ybar, fill=orange!80, draw=orange!80] coordinates {
(0.8,8) (1.8,21) (2.8,11) (3.8,928) (4.8,2097) (5.8,1160)
(6.8,2) (7.8,678) (8.8,514) (9.8,1340) (10.8,13) (11.8,890)
(12.8,2196) (13.8,432) (14.8,487) (15.8,5) (16.8,6) (17.8,8)
(18.8,386) (19.8,0) (20.8,0) (21.8,170) (22.8,196) (23.8,5)
(24.8,213) (25.8,130) (26.8,2) (27.8,8)
};
\addlegendentry{CompGCN}

% UniKER
\addplot[ybar, fill=green!70!black, draw=green!70!black] coordinates {
(1.2,19) (2.2,66) (3.2,37) (4.2,26) (5.2,7) (6.2,35)
(7.2,51) (8.2,31) (9.2,107) (10.2,42) (11.2,63) (12.2,26)
(13.2,10) (14.2,44) (15.2,39) (16.2,20) (17.2,27) (18.2,23)
(19.2,45) (20.2,32) (21.2,52) (22.2,146) (23.2,158)
(24.2,100) (25.2,211) (26.2,141) (27.2,88) (28.2,71)
};
\addlegendentry{UniKER}

\end{axis}
\end{tikzpicture}
\end{subfigure}
\caption{Results on \textsc{Family}. Left: $d_r$ values for 37 rules. Right: $\mathbf{\metrique}_r^{\texttt{hard}}$ values for CompGCN and UniKER on one randomly selected run.}
\label{fig:family}
\end{figure}

\emph{FB15k-237.} 
Figure~\ref{fig:dr_FB15k237} shows that the $d_r$ values on \textsc{FB15k-237} exhibit a much wider range than on \textsc{Family}.
This indicates that many rules extracted or defined for this dataset are only weakly supported by the observed facts. 
For instance, the soft rule $\texttt{area\_admin\_parent}(X,Y) \leftarrow  \texttt{location\_contains}(Y,X)$
has $d_r = 0.88$, meaning that $88\%$ of its evaluable groundings are contradicted in the dataset.
Some rules even reach $d_r=1$, meaning that whenever the body holds in the dataset, the head is violated in the dataset.

Table~\ref{tab:results_FB15k237} compares ExpressGNN, CompGCN, and AnyBURL on 5 runs.
ExpressGNN obtains the best MRR ($0.49$) and Hits@10 ($0.61$). It also obtains the lowest soft-rule score ($\mathbf{\metrique}^{\texttt{soft}}=0.970$). Since $\mathbf{\metrique}^{\texttt{soft}}<1$, ExpressGNN’s predictions are more consistent with the soft rules than the dataset.

The contrast between AnyBURL and CompGCN again illustrates the added value of RVS.
Their Hits@10 and MRR scores are very close (MRR: $0.323$ and $0.335$), making them
difficult to distinguish using predictive accuracy alone.
However, their hard-rule compliance is markedly different:
$\mathbf{\metrique}^{\texttt{hard}}=1.07$ for AnyBURL versus $47.7$ for CompGCN.
Thus, two models with nearly identical ranking performance can have substantially different logical behavior.
This difference is expected given that AnyBURL relies on explicit rule-based inference, whereas CompGCN is an embedding-based neural model.

\begin{figure}[ht]
\centering
\hspace{-3cm}
% ================= LEFT (FIGURE) =================
\begin{minipage}[t]{0.45\textwidth}
\vspace{0pt}
\centering
\begin{tikzpicture}
% Liste des x valeurs
\def\Values{
1.0, 0.7083, 1.0, 1.0, 1.0, 0.9999, 0.0, 1.0, 1.0, 0.0, 1.0, 0.0, 1.0, 1.0, 0.0, 0.0, 1.0, 1.0, 0.0, 1.0
}
\begin{axis}[
    width=5cm,
    height=3.5cm,
    ymin=-0.01,
    ymax=1.11,
    xmin=0.5,
    xmax=20.5,
    xlabel={\parbox{5cm}{Figure \ref{fig:dr_FB15k237}: Values of $d_r$ for 20 sampled rules on \textsc{FB15k-237}.}},
    ylabel={$d_r$},
    xtick={1,...,20},
    xticklabels={
        1,,,,,,,,,
        10,,,,,,,,,,
        20
    },
    ymajorgrids,
    enlargelimits=false,
]
\foreach[count=\i] \v in \Values {
    \pgfmathparse{\v==0 ? 1 : 0}
    \ifnum\pgfmathresult=1
        \addplot[ybar, bar width=3pt, fill=red!60, draw=red!80]
        coordinates {(\i,\v)};
    \else
        \addplot[ybar, bar width=3pt, fill=blue!70, draw=blue!90]
        coordinates {(\i,\v)};
    \fi
}
\end{axis}
\end{tikzpicture}
%\captionof{figure}{}
\phantomcaption
\label{fig:dr_FB15k237}
\end{minipage}
% ================= RIGHT (TABLE) =================
\begin{minipage}[t]{0.5\textwidth}
\vspace{0.3cm}
\centering
\small
\setlength{\tabcolsep}{4pt} % shrink table spacing
\begin{tabular}{|c|c|c|c|c|}
\hline
Method & Hits@10 & MRR & $\mathbf{\metrique}^{\texttt{soft}}$ & $\mathbf{\metrique}^{\texttt{hard}}$ \\ \hline \hline
ExpressGNN & \textbf{0.61 (0.02)} & \textbf{0.49 (0.02)} & \textbf{0.97 (0.00)} & \textbf{5.60 (4.05)} \\ \hline
CompGCN    & 0.51 (0.00) & 0.33 (0.00) & 0.99 (0.00) & 47.7 (7.4) \\ \hline
AnyBURL    & 0.51 (0.00) & 0.32 (0.00) & 0.98 (0.00) & 1.1 (0.6) \\ \hline
\end{tabular}
\captionof{table}{Results on \textsc{FB15k-237} (5 runs). Standard deviations in brackets. Bold = best per metric.}
\label{tab:results_FB15k237}
\end{minipage}
\end{figure}

\emph{DV3F.} On DV3F, $d_r$ varies by property type: House $0.019$ (lowest), Apartment $0.203$, Industrial $0.141$, Outbuilding $0.257$ (Table \ref{tab:dv3f})— the same rule structure has different empirical validity across types, guiding possible rule refinement. GraphSAGE beats Rel-LLM on accuracy ($R^2{=}0.931$ vs.\ $0.915$; MAE $31.9$ vs.\ $40.5$) but Rel-LLM has better aggregate compliance ($0.91$ vs.\ $1.17$). GraphSAGE is more compliant on House ($0.732$ vs.\ $0.915$); Rel-LLM is much more compliant on Outbuilding ($0.857$ vs.\ $1.28$); both are similar on Industrial. The LLM-enhanced model does not improve compliance uniformly — it shifts which regularities are best preserved.

\begin{table}[h]
\centering
\begin{tabular}{lrr|rrrr|r}
\toprule
& $R^2$ & MAE & House & Apartment & Outbuilding & Industrial & RVS agg.\\
\midrule
$d_r$ & & & 0.019 & 0.203 & 0.257 & 0.141\\
GraphSAGE & \textbf{0.931} & \textbf{31.9} & \textbf{0.732} & 1.01 & 1.28 & 0.015 & 1.17\\
Rel-LLM & 0.915 & 40.5 & 0.915 & 1.02 & \textbf{0.857} & 0.015 & \textbf{0.91}\\
\bottomrule
\end{tabular}
\caption{Rule-level $d_r$ and RVS on DV3F. Bold = best per metric.}
\label{tab:dv3f}
\vspace{-0.4cm}
\end{table}

\paragraph{Interpretation.}

Across all three benchmarks, RVS reveals differences between models that are not
captured by conventional predictive-performance metrics. On \textit{Family},
CompGCN achieves better ranking performance than UniKER but produces approximately
$7\times$ more hard-rule violations. On \textsc{FB15k-237}, AnyBURL and CompGCN
have nearly identical MRR, yet their hard-rule RVS differs by more than an order of
magnitude. On DV3F, GraphSAGE achieves better $R^2$ and MAE, whereas Rel-LLM
achieves better aggregate logical compliance.

These results show that predictive performance and logical compliance capture
distinct aspects of model behavior. RVS additionally provides rule-level diagnostics,
revealing which constraints account for differences between models and whether
compliance is uniform across rules.

The same evaluation framework applies across tasks and data models, including link
prediction and relational regression, and both knowledge graphs and relational
databases. Overall, RVS complements conventional predictive-performance metrics at
three levels: model comparison, dataset auditing, and rule-level diagnosis.

\section{Related Work}

Task-specific accuracy metrics (Hits@k, MRR, R\textsuperscript{2}, MAE, accuracy, precision, AuROC/AuPRC \cite{r2-metric,chicco2021coefficient,mae-metric,precision-metric,fawcett2006introduction,saito2015precision,auc-neurips24}) measure prediction–ground-truth agreement but say nothing about logical constraints, and none is reported alongside a compliance figure in current benchmarks — precisely the gap RVS fills.

\textbf{Semantic loss approaches.} A separate line of work injects symbolic knowledge into neural systems as a \emph{training-time loss}, not an \emph{evaluation metric}. Several works like \citet{semanticloss,badreddine2022logic,zhang2023improved,ahmed2022neuro,ahmed2023pseudo,arrotta2024semantic} define a semantic loss over propositional-logic constraints, added to the task loss during training, computed from output probability distributions rather than discrete predictions checked against a relational rule set. Trying to adapt such a signal in order to obtain a logic-based metric might be tempting at first sight, but this is far from trivial and requires  significant work as it raises several important issues. Semantic loss \citet{semanticloss} is the seminal work that introduced this idea. First, it is a single differentiable scalar tuned for optimization, not a ratio of counts with a dataset baseline (P2, P5) — a loss of $0.31$ says nothing about whether that is good or bad relative to the data, unlike $\mathrm{RVS}^{\mathrm{soft}}_r$. Second, it does not take into account the hard/soft distinction (P2) or rule importance (P4, P5). 
Last, and most importantly, it is designed to be computed on the output of a neural network, not to be model-agnostic (P7), so this would require a complete redesign in order to adapt it to relational rules with variables and joins (P6).

\textbf{Vision tasks.}
Similar ideas have been developed for convolutional neural networks (CNNs) \cite{schwalbe2022enabling}, but the framework is limited to vision tasks, and not applicable to our relational setting.

\textbf{Semantic evaluation metrics for KGs.} Closer in spirit, since it \emph{is} an evaluation metric, is Sem@k \cite{semk}: for each link-prediction query, it checks whether top-$k$ predicted entities respect the domain/range of the relation, reporting the fraction of semantically valid predictions. Sem@k targets one task (link prediction), one data model (knowledge graphs), and one rule type (domain/range), with no mechanism for rule importance or a hard/soft split. RVS generalizes across rules, tasks, and data models including relational databases (P6), supports user- and dataset-derived rule importance (P4, P5), and separates hard from soft rules (P2). Sem@1 is in fact a special case of RVS under two rules constraining subject/object types to a relation's domain/range.

\section{Conclusion and Future Work}

We introduced \metrique{}, an evaluation metric for measuring compliance with logical rules independently of predictive performance. Its design is guided by seven properties, including the distinction between hard and soft rules, direct interpretability, user- and dataset-derived rule importance, relational scope, and model-agnostic computability. \metrique{} provides rule-level and aggregate diagnostics, enabling logical compliance to be evaluated alongside conventional predictive-performance metrics. Experiments across knowledge-graph link prediction and relational regression show that models with comparable predictive performance can exhibit substantially different levels of logical compliance, and that better predictive performance does not necessarily imply better compliance. Beyond model comparison, the dataset contradiction rate $d_r$ provides a model-independent diagnostic of how well individual rules are supported by the observed data. %
Taken together, these results position \metrique{} as a first step toward making logical compliance an explicit component of model evaluation. Its current formulation evaluates predictions individually, enabling prediction-level attribution but not capturing violations that arise only through interactions among multiple predictions. Extending \metrique{} to jointly evaluate sets of predictions is therefore a natural direction for future work. 
More broadly, our findings highlight the importance of evaluating logical compliance as a distinct dimension of model behavior, and suggest that such measures should increasingly complement predictive-performance metrics in AI evaluation protocols.

\begin{ack}
This work was supported by the NSRL Chair of the MIAI Cluster,
funded by the French government through the France 2030
investment program, managed by the French National Research
Agency (ANR), under grant number ANR-23-IACL-0006.
\end{ack}

%%%%%%%%%%%%%%%%%%%%%%%%%%%%%%%%%%%%%%%%%%%%%%%%%%%%%%%%

\bibliographystyle{plainnat}
\bibliography{references}

%%%%%%%%%%%%%%%%%%%%%%%%%%%%%%%%%%%%%%%%%%%%%%%%%%%%%%%%%%%%

\appendix

\section{Hyperparameters}\label{apd:hyperparameters}

All experiments are conducted on a Linux server equipped with an Intel Xeon Silver 4210R CPU (10 cores, 20 threads, 2.40 GHz), 180 GB of RAM, and an NVIDIA RTX A6000 GPU with 49 GB of VRAM. 
The full implementation details, configuration files and results are publicly available in our GitHub repository\footnote{https://anonymous.4open.science/r/Rule-Violation-Score-585C}.
Throughout all experiments, the importance weight $\sigma_r$ is set to $1$ for every rule $r \in R_{\mathrm{hard}}$.

\subsection{Inductive model: UniKER}

The UniKER model is trained in TransE mode with a fixed configuration across experiments.

\begin{description}
    
    \item [Mode:] TransE

    \item [Iterations:] 8
  
    \item [Noise threshold:] 0.0

    \item [Top-k threshold:] 0.2

    \item [Init flag:] 0 (default)
\end{description}

\subsection{Rule-based model: AnyBURL}

The AnyBURL model is trained using identical hyperparameters for both \textsc{Family} and \textsc{FB15k-237}.

\begin{description}
    \item [WORKER\_THREADS:] 7
    \item [POLICY:] 2
    \item [REWARD:] 5
    \item [EPSILON:] 0.1
    \item [THRESHOLD\_CORRECT\_PREDICTIONS:] 2
    \item [THRESHOLD\_CONFIDENCE:] 0.0001
    \item [ZERO\_RULES\_ACTIVE:] false
\end{description}

\subsection{Embedding model: CompGCN}

\paragraph{Family setting:}
\begin{description}
\item[name:] best\_model
\item[score\_func:] conve
\item[opn:] corr
\end{description}
\paragraph{FB15k-237 setting:}
\begin{description}
\item[score\_func:] transe
\item[opn:] sub
\item[gamma:] 9
\item[hid\_drop:] 0.1
\item[init\_dim:] 200
\end{description}

\subsection{GNN model: ExpressGNN (FB15k-237)}

\begin{description}
\item[slice\_dim:] 16
\item[batchsize:] 16
\item[use\_gcn:] 1
\item[num\_hops:] 1
\item[embedding\_size:] 128
\item[gcn\_free\_size:] 127
\item[patience:] 20
\item[lr\_decay\_patience:] 100
\item[entropy\_temp:] 1
\end{description}

\subsection{GraphSAGE (DV3F)}

GraphSAGE is used with the default settings from the original implementation.

\subsection{Rel-LLM (DV3F)}

\begin{description}
\item[epochs:] 15
\item[batch\_size:] 128
\item[learning rate:] 2e-4
\item[dropout:] 0.1
\item[LLM:] frozen
\item[output layer:] MLP
\end{description}

\subsection{Evaluation infrastructure}

All evaluation metrics are computed using a PostgreSQL database running inside a Docker container. Predictions are generated in Python and automatically translated into SQL queries.
The metric is computed in 10 seconds on the \textsc{Family} dataset, 5 minutes on \textsc{FB15k-237}, and 10 minutes on DV3F.

\section{Practical Computation of \metrique}\label{apd:computation}

We describe how we compute \metrique{} in practice for any dataset, prediction algorithm, and rule set.
Computing \metrique{} requires two preliminary steps: the algorithm must first be run on the test set to obtain predictions, and both the dataset and the predictions must then be stored in a database, as described below.

\paragraph{Database schema.}
We create one table per predicate symbol appearing in any rule $r$.
We distinguish two settings depending on the data model.

\begin{itemize}
    \item \textbf{Knowledge graph.}
    The dataset is a set of triples $(s, r, o)$, where $s$ and $o$ are
    entities and $r$ is a relation.
    Each relation $r$ becomes a dedicated table with columns
    \texttt{subject}, \texttt{object},
    \texttt{tag}, and \texttt{model}.
    Observed facts are inserted with $\mathtt{tag} = 0$ and
    \texttt{model} is left \texttt{NULL}.
    Predictions are inserted with $\mathtt{tag} = 1$, \texttt{model} set to
    the algorithm identifier.
    The facts in the test data are not inserted into these tables.

    \item \textbf{Relational database.}
    The original schema is preserved as-is.
    In the table containing the test facts, we add a column \texttt{tag} set to $2$ for the test facts and $0$ for observed facts.
    As before, test data are not used to compute RVS.
    Then, to distinguish the predictions made by the algorithm, we create a new table with columns \texttt{id}, \texttt{prediction}, \texttt{tag}, and \texttt{model}, filled using the same convention as above.
\end{itemize}

Table~\ref{tab:example-aunt} shows a sample of the \texttt{aunt} table
for the \textsc{Family} dataset, illustrating the coexistence of observed
facts and predictions from a single model in the same table.

\begin{table}[h]
    \centering
    \caption{Sample of the \texttt{aunt} table for the \textsc{Family}
    dataset.
    Rows with $\mathtt{tag} = 0$ are observed facts;
    rows with $\mathtt{tag} = 1$ are model predictions.}
    \label{tab:example-aunt}
    \begin{tabular}{rrrr}
        \toprule
        \texttt{subject} & \texttt{object} & \texttt{tag} & \texttt{model} \\
        \midrule
        1369 & 1287 & 0 & NULL \\
        1682 &  512 & 0 & NULL \\
        1283 & 1271 & 0 & NULL \\
        2079 & 2088 & 1 & \texttt{compgcn} \\
        2082 & 2086 & 1 & \texttt{compgcn} \\
        2099 & 2089 & 1 & \texttt{compgcn} \\
        \bottomrule
    \end{tabular}
\end{table}

\paragraph{RVS computation with SQL.}
For each rule $r$, the quantities $d_r$ and $p_r$ are computed using SQL
queries.
We automated the generation of these SQL queries for Horn rules. In the case of a knowledge graph, all facts are of the form \texttt{rel(subject, object)}.
One query is shown in the following example.
For other types of rules, we need to define SQL queries manually.

\paragraph{Example query}
Consider the rule
$r: \texttt{aunt}(X, Z) \leftarrow \texttt{sister}(X, Y) \wedge
\texttt{father}(Y, Z)$
on the \textsc{Family} dataset.
To count the confirmed groundings
$C_r^D =
\{\theta \in B_r^D \mid \psi_r\theta \texttt{ is confirmed in } D \}
$ 
needed for the computation of $d_r$, we issue:

\begin{lstlisting}[language=SQL]
SELECT COUNT(*)
FROM   aunt   t0
JOIN   sister t1 ON t0.subject = t1.subject
JOIN   father t2 ON t1.object  = t2.subject
               AND t0.object   = t2.object
WHERE  t0.model IS NULL
  AND  t1.model IS NULL
  AND  t2.model IS NULL
  AND  t0.subject <> t1.object
  AND  t0.subject <> t0.object
  AND  t1.object  <> t0.object;
\end{lstlisting}
This corresponds to evaluating every grounding of $\texttt{aunt}(x,z) \land \texttt{sister}(x,y) \land \texttt{father}(y,z)$.

The three \texttt{model IS NULL} filters restrict the query to observed
facts.
The three inequality conditions enforce that the variables $X$, $Y$, $Z$
are mapped to distinct constants.
To count violating groundings (the numerator of $d_r$), we adopt the open-world assumption. Thus, we first define a set of relations incompatible with \texttt{aunt}: \texttt{brother}, \texttt{sister}, \texttt{father}, ...
Then, for each \texttt{incomp\_rel} in this set, we run the above SQL query by substituting \texttt{aunt} with \texttt{incomp\_rel}, and sum the resulting counts.

To compute $p_r$ for a model \texttt{m}, the same query structure is reused with two modifications.
First, the condition
\begin{lstlisting}[language=SQL]
    (t0.tag + t1.tag + t2.tag) = 1
\end{lstlisting}
is added to ensure that exactly one atom in each grounding originates from the predictions of model \texttt{m}, while the remaining atoms come from observed facts.
This makes every prediction contribute independently to the count.
Second, the filter 
\begin{lstlisting}[language=SQL]
    t0.model = 'm' OR t1.model = 'm' OR t2.model = 'm'
\end{lstlisting}
is added to identify the model.

Given $d_r$ and $p_r$, the final metrics $\textbf{RVS}_r^{hard}$ and $\textbf{RVS}_r^{soft}$ are obtained as described in Section~\ref{sec:dr_pr_score}.
These are then aggregated to compute $\textbf{RVS}^{\mathrm{hard}}$ and $\textbf{RVS}^{\mathrm{soft}}$, as defined in
Section~\ref{sec:global_metric_aggregation}.

\section{Incompatibility Constraints on \textsc{FB15k-237}}\label{apd:incompatibility}

To define the set of refutation rules $R_{\texttt{ref}}$, it is necessary to identify pairs of relations that are incompatible within the \textsc{FB15k-237} dataset.
We adopt the open-world assumption (OWA), under which absence from the knowledge graph does not imply falsity.
Nevertheless, for the purpose of deriving incompatibility constraints, we introduce an additional assumption: if two relations have never been observed to hold simultaneously for the same pair of entities in the dataset, they are considered incompatible.
While this assumption enables automatic extraction of such constraints, these may also be specified manually when needed.

For each relation \( rel_i \), we define the set of incompatible relations as
\[
\texttt{all\_incompat\_rels}(rel_i)
=
\left\{
rel_k \in \texttt{all\_relation}
\;\middle|\;
\nexists (A,B)\in \mathcal{D}^2,\;
rel_i(A,B)\land rel_k(A,B)
\right\}.
\]
To compute this set, we first identify, for every relation \( rel_i \), the set of relations that are compatible with it.
Starting from an empty compatibility set, we iterate over all facts in the dataset.
Whenever both \( rel_i(A,B) \) and \( rel_k(A,B) \) are observed in the dataset for any pair of entities \( (A,B) \), the relation \( rel_k \) is marked as compatible with \( rel_i \).

After processing the entire dataset, any relation that has not been observed to co-occur with \( rel_i \) for any entity pair is deemed incompatible with \( rel_i \).

These incompatibility constraints are then used to derive refutation rules of the form
\[
\bot \leftarrow rel_i(X,Y) \land rel_k(X,Y),
\]
for every $rel_i \in \texttt{all\_relations}$ and \( rel_k \in \texttt{all\_incompat\_rels}(rel_i) \).

Formally:
\[
R_{\texttt{ref}} = \{
    \bot \leftarrow rel_i(X,Y) \land rel_k(X,Y)
    \mid
    rel_i \in \texttt{all\_relations} \texttt{ and } rel_k \in \texttt{all\_incompat\_rels}(rel_i)\}
\]

%%%%%%%%%%%%%%%%%%%%%%%%%%%%%%%%%%%%%%%%%%%%%%%%%%%%%%%%%%%%

%\newpage
%\input{checklist.tex}

\end{document}